\documentclass[letterpaper, 10 pt, conference]{ieeeconf}  

\IEEEoverridecommandlockouts                              

\usepackage{xcolor}
\usepackage{color}
\usepackage{amssymb}
\usepackage{amstext}
\usepackage{algorithm}
\usepackage{algorithmicx}
\usepackage{algpseudocode}
\usepackage{booktabs}
\usepackage{multirow}
\usepackage{graphicx}
\usepackage{amsmath}
\usepackage[caption=false,font=normalsize,labelfont=sf,textfont=sf]{subfig}

\RequirePackage[normalem]{ulem} 
\RequirePackage{xcolor}

\begin{document}
\title{\LARGE \bf
Learning Category-Level Generalizable Object Manipulation Policy via\\
Generative Adversarial Self-Imitation Learning from Demonstrations
}

\author{{Hao Shen$^{1*}$, Weikang Wan$^{1*}$ and He Wang$^{1\dagger}$}
\thanks{*: contribute equally, $\dagger$: corresponding author.}
\thanks{$^{1}$ Hao Shen ({\tt\small shenhaosim@pku.edu.cn}) and Weikang Wan ({\tt\small wwk@pku.edu.cn}) are with Peking University, Beijing, China. He Wang ({\tt\small hewang@pku.edu.cn}) is with Faculty of Center on Frontiers of Computing Studies (CFCS), Peking University, Beijing, China.}}%


\maketitle
\thispagestyle{empty}
\pagestyle{empty}

\begin{abstract}
Generalizable object manipulation skills are critical for intelligent and multi-functional robots to work in real-world complex scenes. 
Despite the recent progress in reinforcement learning, it is still very challenging to learn a generalizable manipulation policy that can handle a category of geometrically diverse articulated objects.
In this work, we tackle this category-level object manipulation policy learning problem via imitation learning in a task-agnostic manner, where we assume no handcrafted dense rewards but only a terminal reward.
Given this novel and challenging generalizable policy learning problem, we identify several key issues that can fail the previous imitation learning algorithms and hinder the generalization to unseen instances. 
We then propose several general but critical techniques, including generative adversarial self-imitation learning from demonstrations, progressive growing of discriminator, and instance-balancing for expert buffer, that accurately pinpoints and tackles these issues and can benefit category-level manipulation policy learning regardless of the tasks.
Our experiments on ManiSkill benchmarks demonstrate a remarkable improvement on all tasks and our ablation studies further validate the contribution of each proposed technique.

\end{abstract}
\section{INTRODUCTION}
\vspace{-1mm}
Learning human-level object manipulation skills is highly demanding for robots, which can automate repetitive works in our real-world complex scenes and lead to revolutionary applications, \textit{e.g.} home robots, that may impose huge impacts on human society. A crucial and valuable feature of human manipulation skills is the remarkable generalization ability -- for certain tasks (\textit{e.g.} open the door), we can successfully manipulate different kinds of object instances (that contain a door), regardless of the large variations in their geometry and topology. 
With the emerging deep reinforcement learning (RL) and imitation learning (IL) techniques, we have recently observed great progress in generalizable robot learning on basic tasks that only involve rigid objects with simple manipulation skills, \textit{e.g.} grasping\cite{levine2015learning,mahler2017dex}, planar pushing\cite{yu2016more}, pick and place\cite{zeng2021transporter}, etc. However, learning generalizable policy for more complex tasks that involve articulated objects, \textit{e.g.} open drawers and doors, are still highly under-explored.

Recently, to facilitate the research and benchmark the progress in learning generalizable object manipulation skills on articulated objects, Mu \textit{et al.}~\cite{mu2021maniskill} initiated SAPIEN open-source manipulation skill (ManiSkill) challenge. The challenge introduces four tasks with each aimed at manipulating one category of articulated objects using robots with either a single arm or dual arms. Based on SAPIEN environment and PartNet-mobilty dataset\cite{xiang2020sapien}, the benchmark split a category of objects into non-overlapped training and testing instances and the learned policy is expected to work on both seen instances as well as novel object instances at test time.
As shown in ~\cite{mu2021maniskill}, learning such category-level generalizable policy is very challenging for RL-based approaches. Naively adopting state-of-the-art RL methods on category-level generalizable policy learning achieves almost zero success rate. To ease the learning, ManiSkill provides many successful demonstrations on training object instances and thus enables learning from demonstrations.

In this work, we thus tackle the problem of learning category-level generalizable object manipulation policy via imitation learning in a task-agnostic manner.
Although, for each individual task, conducting reward engineering that incorporates rich task and object category priors into a handcrafted dense reward does help, we aim to find a more general and universal approach that assumes no handcrafted rewards but only a terminal one, which will benefit all kinds of category-level manipulation tasks and is complimentary to task-specific reward design.

With a sufficient amount of demonstrations, we couple the popular IL method, Generative Adversarial Imitation Learning (GAIL) ~\cite{ho2016generative}, with the prevailing model-free off-policy RL algorithm, Soft Actor-Critic (SAC)~\cite{haarnoja2018soft} and achieve non-zero success rates on both training and test instances. However, this baseline is far from being satisfactory and is not tailored for category-level generalizable policy learning.

Through in-depth investigation, we identify several key problems that hinder the generalizable policy learning: 1) to handle many training instances altogether, the policy network is 
more prone to fail under the GAIL's setting, since the policy network gets a harder task and thus is easily overpowered by the discriminator, whose reward (expert reward) may diminish to zero and further stop the imitation learning. 2) the demonstrations of different object instances may be from different strategies, which is difficult for a single policy network to imitate, given its limited capacity and intrinsic continuity in neural networks; 3) even the policy learning is successful on some training instances, the learned policy can still be highly biased such that it can only handle a certain type of instances and doesn't generalize to novel instances.

To mitigate the aforementioned issues, we propose several important extensions to the baseline. Our technical contribution includes: propose to progressively grow the discriminator of GAIL to mitigate issue 1; propose Generative Adversarial Self-Imitation Learning from Demonstrations to mitigate issue 2; and propose category-level instance balancing expert buffer to mitigate issue 3.

Our proposed method achieves remarkable improvements on success rates, outperforming the GAIL+SAC baseline by \textbf{13\%} and \textbf{18\%} averaged across four tasks on training and validation sets, respectively. On the ``no external annotation" track of ManiSkill Challenge 2021\footnote{https://sapien.ucsd.edu/challenges/maniskill2021/} that allows interaction,
dense reward but no further annotations, our method ranks first place when further coupling with dense rewards.

\section{RELATED WORK}
\vspace{-2mm}
\subsection{Learning Generalizable Manipulation Skills}
Generalization is essential for future robot applications. The robots are expected to be functional in a variety of real-world environments. 
For simple motion tasks like pushing and lifting, the variation in objects seldom constraints the performance of algorithms. However, for dealing with articulated objects whose each part has unique dynamics and function, generalization becomes challenging.
Previous works focus to identify key parts or extract features of articulations as representations~\cite{mittal2021articulated,arduengo2021robust,devin2018deep} to enable generalized manipulation on different instances. These methods are based on visual information including key location identification, pose estimation, or pretrained attention models. There are also control-based methods using model prediction and generative planning methods~\cite{abbatematteo2019learning,jain2020learning} to achieve robust and adaptive control on both seen and novel objects.

\subsection{Imitation Learning from Demonstrations} Imitation learning (IL) techniques aim to mimic expert behaviors in a certain task. When expert demonstrations of the tasks are given, a task-specific reward function becomes optional. The ordinary way to perform imitation learning includes behavior cloning~\cite{bain1995framework} and inverse reinforcement learning~\cite{ng2000algorithms,russell1998learning}.
In recent years, a new IL method has emerged, namely, generative adversarial imitation learning (GAIL)~\cite{ho2016generative} which has achieved state-of-the-art performance on numerous complex robotics tasks~\cite{merel2017learning,wang2017robust}. This method incorporates RL into the GAN framework. The generator which is the policy of the RL part generates demonstrations in the environment, and the discriminator is used to discriminate between the generated and expert demonstrations, such that the policy of RL converges to the expert policy. To improve the sample efficiency, there have been numerous studies leveraging off-policy RL algorithms rather than original on-policy RL algorithms for policy generation~\cite{kostrikov2018discriminator}.
In addition, GASIL~\cite{guo2018generative} combines the idea of self-imitation~\cite{oh2018self} with GAIL to realize imitation learning without demonstration data. Moreover, SILfD~\cite{pshikhachev2021self} uses expert demonstration data to help self-imitation learning.

However, using imitation learning to train an agent that utilizes expert demonstrations effectively and generalizes to different unseen objects still remains challenging.

\subsection{3D Articulated Object Manipulation}
3D articulated object manipulation is at the nexus of computer vision and robotics and has been an attractive topic in research. 
Previous works utilize the learning methods to get the vision knowledge (such as articulated part configurations, link poses, joint parameters, and state information) of the environment from the 3D input in order to build a bridge between vision and robotics and help the downstream robotic tasks~\cite{wang2019shape2motion,li2020category,zeng2020visual,mu2021sdf}.  Other works like~\cite{abbatematteo2019learning} estimate kinematic structures to predict the shape and kinematic model of an object from depth sensor data.
On the other hand, there have also been impressive works about various robotic planning, trajectory optimization, and control methods on articulated object manipulation~\cite{peterson2000high,chitta2010planning} like door-opening. These methods typically assume perfectly known geometry with objects pre-fixed to the robot.
Recent works have demonstrated successful systems which leverage the vision learning methods and the robotic planning methods to manipulate 3D articulated objects with the learned visual knowledge~\cite{gadre2021act,harada2019service,schmid2008opening}.
These works typically start with the vision methods such as pose estimation, object detection, and part segmentation to get the visual knowledge of the environment and compute a motion trajectory. 
However, this explicit standardized visual knowledge may be insufficient for the downstream robotic tasks if the diverse articulated objects have complex topological and geometric variations, such as different numbers and shapes of doors and drawers on different shapes of cabinets. In this way, inaccuracy is caused by visual perception.

\section{PROBLEM FORMULATION}
\vspace{-1mm}
Our problem concerns the policy learning for category-level generalizable articulated object manipulation, which requires robots to manipulate a variety of articulated objects from one object category for a specific task, e.g. opening different types of drawers. The tasks and the environments come from the ManiSkill Challenges~\cite{mu2021maniskill}. Our environment thus comprises a robot and an articulated object and multiple depth cameras, from which we can obtain a fused point cloud observation of the environment. The goal is specified by instance mask and part mask of the point cloud, for example, the target drawer and its corresponding handle are labeled in the opening drawer task, and the task is considered successful when the target drawer is opened to a certain extent and remain static. Assuming the robot state is always known, our state $S_t$ thus comprises a colored point cloud with per-point part labels and the robot state. 

In this work, we tackle the problem under the setting of imitation learning, following the setting of the ``no external annotation'' track in ManiSkill Challenges, where several successful demonstrations are available for each training object instance but no external annotations are allowed. We further get rid of any hand-crafted task-specific reward functions and only assume a terminal reward $r_t$ (1 if success, otherwise 0), aiming to learn a category-level generalizable manipulation policy via imitation learning from demonstrations in a task-agnostic manner. 
The goal is to learn a policy that can handle both training and unseen novel instances.

\section{METHOD}
\vspace{-1mm}

\subsection{GAIL with SAC}
\label{sec:gail}
\begin{figure}[!t]
\vspace{+2mm}
\centering
\includegraphics[width=3.2in]{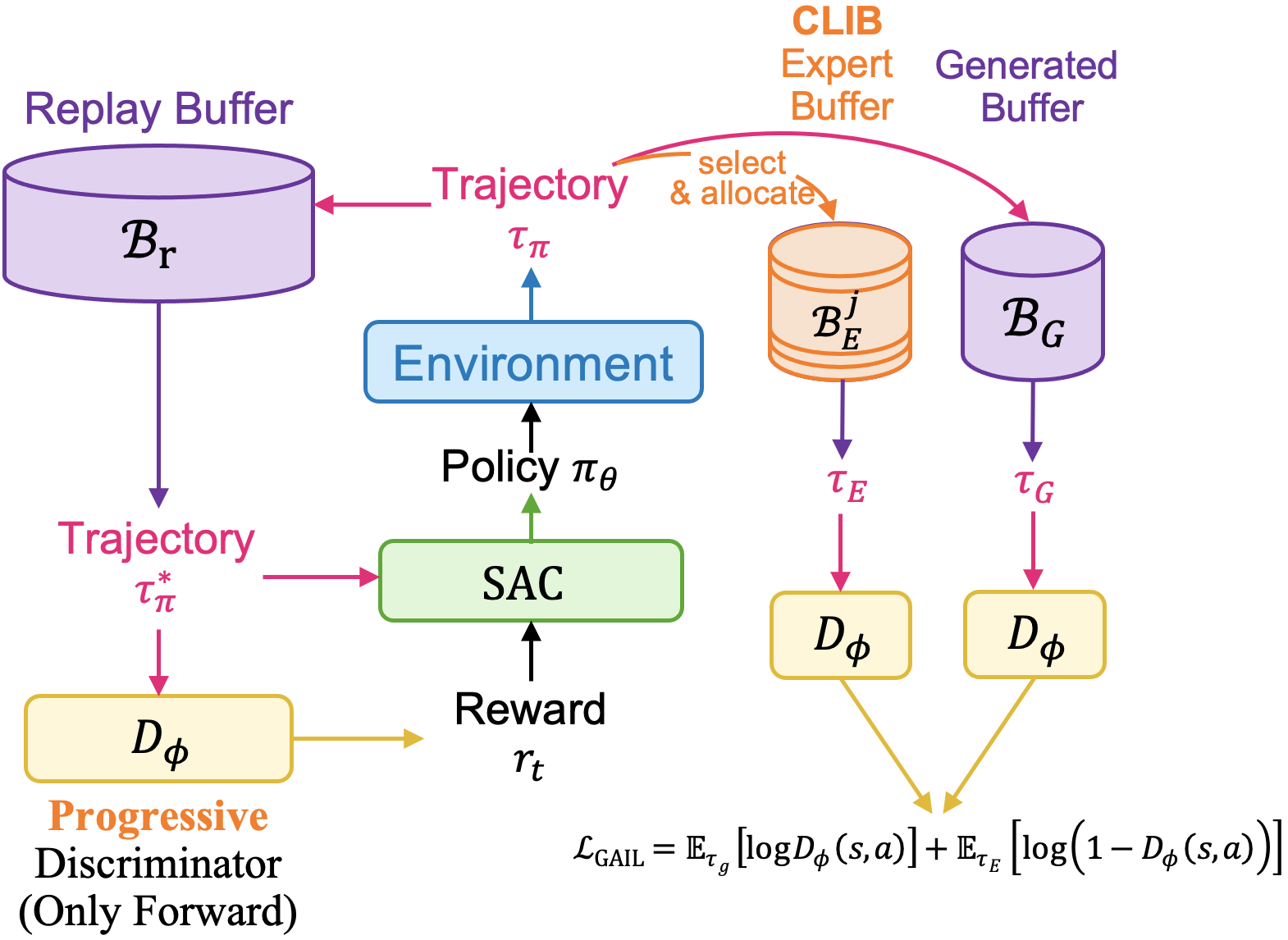}
\caption{\textbf{Pipeline Overview}. On top of Generative Adversarial Imitation Learning, we introduce Category-Level Instance-Balancing (CLIB) Expert Buffer, which both includes expert demonstrations and successful trajectories and maintain a balance between different instances of objects. Besides, we modified the discriminator's structure to make it progressive as training goes. (The orange parts indicate our proposed techniques.) }
\label{fig:pipeline}
\vspace{-5mm}
\end{figure}

Our baseline implements GAIL~\cite{ho2016generative} to utilize demonstration data provided by ManiSkill Challenge \cite{mu2021maniskill}. These expert demonstrations $\{\tau_{E}\}$ are generated by training per-instance manipulation policy using a popular off-policy model-free reinforcement learning method, Soft Actor-Critic (SAC) algorithm~\cite{haarnoja2018soft}, with task-specific handcrafted dense rewards. Without using these task-specific rewards, we thus couple GAIL with SAC to provide dense rewards. In this baseline, GAIL utilizes an adversarial trajectory generation scheme that comprises a generator, which is simply the policy network of SAC, and a discriminator $D_{\phi}$, which trains to distinguish between expert trajectories $\tau_{E}$ and generated trajectories 
$\tau_{G}$
and whose outputs can provide dense rewards 
$r_t$
to SAC. This off-policy imitation learning framework contains three buffers: replay buffer $\mathcal{B}_r$, generated buffer
$B_{G}$
(generated trajectories), and expert buffer $\mathcal{B}_E$. 
As for our whole pipeline, we further make the discriminator $D_{\phi}$ progressive, use self imitation learning to update the expert buffer, and use a Category-Level Instance-Balancing (CLIB) Expert Buffer $\mathcal{B}_E^j$. See Algorithm \ref{alg:1} for the whole pipeline. 
The details are shown in Fig. \ref{fig:pipeline}.

\begin{algorithm}[!b]\small
  \centering
  \label{alg:1}
  \caption{Category-Level Generative Adversarial Self-Imitation Learning from demonstrations (underlined part indicates our techniques added to GAIL with SAC)}
  \begin{algorithmic}[1]
    \State Initialize each instance expert buffer with expert demonstration trajectories $\mathcal{B}_E^j \leftarrow \tau_E^j$, where $j=1,2,...,n$ and $n$ indicates the total number of instances.
    \For{each iteration}
      \State Initialize Generated Buffer $\mathcal{B}_G \leftarrow \emptyset$
      \State Use SAC to generate trajectories $\tau_\pi \sim \pi_\theta$
      \State Set Generated Buffer $\mathcal{B}_G \leftarrow \tau_\pi$
      \State Update the replay buffer of SAC $\mathcal{B}_r$ using $\tau_\pi$
      \State \underline{Select successful trajectories $\Tilde{\tau}_\pi^j \subset \tau_\pi$ and update each}  \underline{instance expert buffer $\mathcal{B}_E^j$ using $\Tilde{\tau}_\pi^j$ for $j=1,2,...,n$}
      
      \State Sample trajectories from replay buffer
      $$\tau_\pi^* = \{(S_t, a_t, r_t, S_{t+1})\} \sim \mathcal{B}_r$$\qquad where reward $r_t$ from \underline{progressive discriminator} $D_\phi$ is: $r_t = -\log D_\phi(S_t,a_t)$
      \State Update SAC parameters using $\tau_\pi^*$
      \For{each discriminator update}
        \State Sample expert trajectories \underline{$\tau_E^{j*} \sim \mathcal{B}_E^j$, $\tau_E=\bigcup_{j=1}^{n}\tau_E^{j*}$} and generated trajectories $\tau_G \sim \mathcal{B}_G$
        \State Update discriminator parameter using $\tau_E$ and $\tau_G$:
        $$\nabla_\phi \mathcal{L}_{\text{GAIL}}=\mathbb{E}_{\tau_G}[\nabla_\phi\log D_\phi(s,a)]+\mathbb{E}_{\tau_E}[\nabla_\phi\log(1- D_\phi(s,a))]$$
      \EndFor
    \EndFor
  \end{algorithmic}
\vspace{-1mm}
\end{algorithm}

\begin{figure*}[!t]
\vspace{+2mm}
\centering
\includegraphics[width=5.6in]{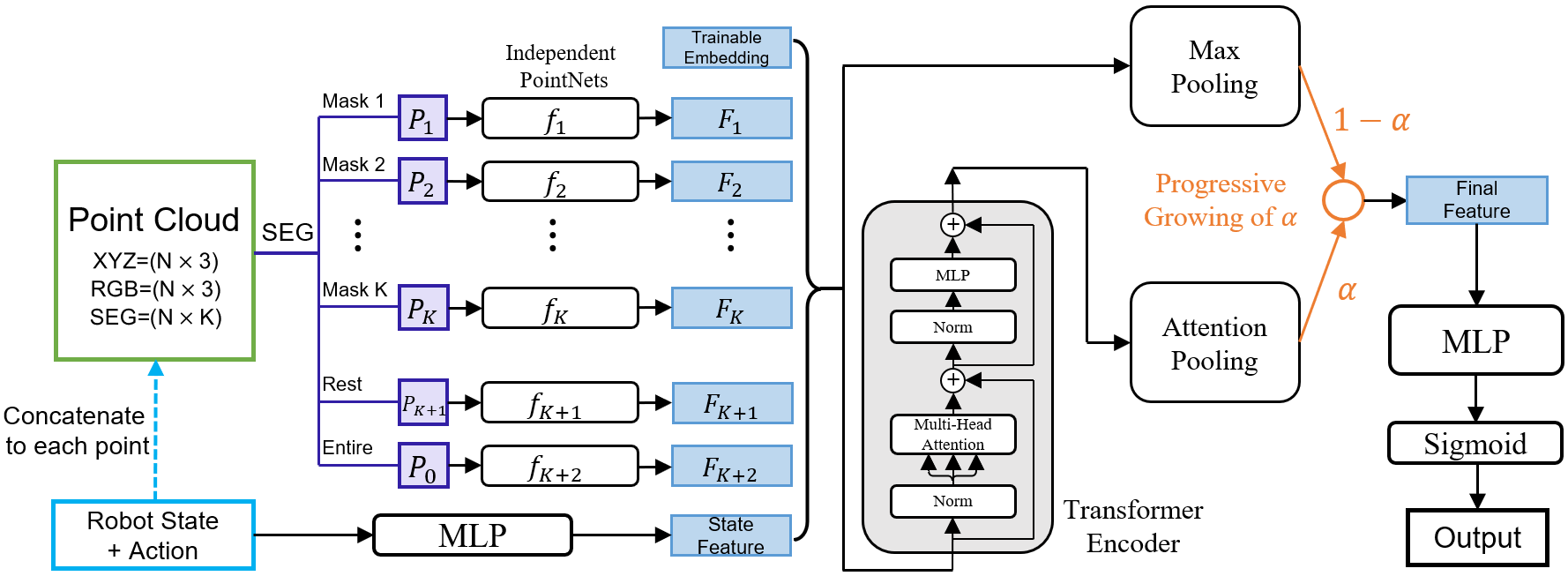}
\caption{\textbf{The progressive structure of discriminator network.} The input contains the point cloud, robot state, and action. The robot state and action are concatenated to a vector in the figure. The output of the network is a scalar ranging from 0 to 1. During training, the latter part of the network progressively grows from max pooling to attention pooling.}
\label{fig:progressive}
\vspace{-4mm}
\end{figure*}

\vspace{-1mm}
\subsection{Progressively Growing the Discriminator of GAIL}
\label{sec:progressive}
In GAIL, the reward (which we refer as expert reward) is set to be $r'_t = -\log D_\phi(S_t,a_t)$. Whether the policy network can constantly receive positive reward signals is crucial to the success of imitation learning. 
Due to the need to handle many training instances with large variations, the policy network
is more prone to fail under the GAIL's setting
at the very beginning and can easily be over-powered by the discriminator.  In other words, 
the discriminator is usually too strong and can easily tell the generated trajectories from the expert demonstration trajectories soon after the training starts. 
This will cause the reward from GAIL to quickly vanish to zero, leading to failures in imitation learning. 

Inspired by Progressive Growing GAN (PG-GAN) \cite{karras2018progressive} and Curriculum GAN \cite{sharma2018improved}, we propose to progressively grow the architecture of discriminator during the training, to alleviate the imbalance. 
More specifically, we let the discriminator evolve from a simple initial architecture to a more complex one. 
Here we adopt the baseline PointNet architecture proposed by \cite{mu2021maniskill} as our initial architecture while using the PointNet + Transformer architecture in \cite{mu2021maniskill} as our final architecture. 
As shown in Fig. \ref{fig:progressive}, 
both the initial and final architectures share the same backbone, which comprises K+2 independent PointNet $f_i$ ($i = 1, 2,..., K+2$), where $K$ is the number of part masks defined by the task, $f_i (i\leq K)$ extract the feature of part point cloud $P_i$, $f_{K+1}$ extracts the feature of the rest points, and $f_{K+2}$ extracts the feature of all points $P_0$ together. Taking these K+2 features along with a trainable embedding vector (serving as a bias for the task) and a processed robot state 
as inputs, our initial architecture simply performs a max-pooling along 
the 
K+4 features to generate the final feature, while our final architectures further leverage a Transformer layer to process the features and then perform an attention pooling.
When progressive growing is triggered, 
a parameter $\alpha$ is introduced to linearly interpolate between the output of max pooling and attention pooling (Fig. \ref{fig:progressive}). The network also includes stabilizing phases before and after the growth of $\alpha$. At the full growth of the progressive network, the structure is the same as PointNet + Transformer structure in \cite{mu2021maniskill}. In the end, the global feature goes through an MLP and Sigmoid layer to form the output of the discriminator.

The underlying philosophy of our method is indeed curriculum training and echoes Curriculum GAN, which gradually increases the sophistication of the discriminator to make the generator learn smoothly. In our case, the discriminative power of the discriminator is gradually increased and thus the discriminator is more generous with its reward at the beginning of the training than later, allowing the policy network to slowly warm-up. Note that we don't evolve the architecture of the generator (policy network) based on the following two reasons: 1) the output of our policy network is always in the same action space without any change, which is different from PG-GAN\cite{karras2018progressive} that increases the generation resolution from coarse to fine; 2) since the issue of GAIL comes from the overpowered discriminator, we want the policy network to be as strong as possible from the very beginning. 

\subsection{Generative Adversarial Self-Imitation Learning from Demonstrations}
\label{sec:SILfD}
\begin{figure}[!t]
\centering
\subfloat[]{\includegraphics[width=1.7in]{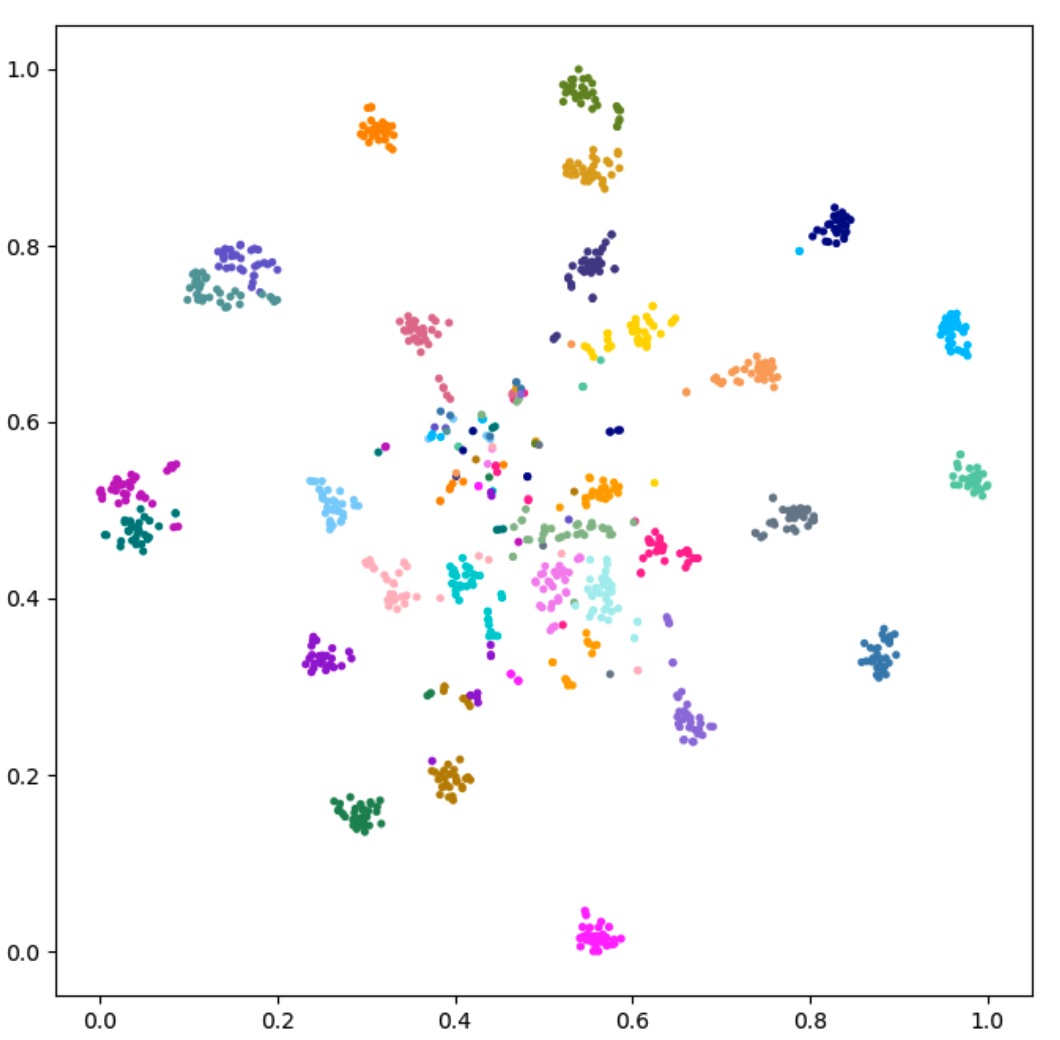}%
\label{fig_first_case}}
\hfil
\subfloat[]{\includegraphics[width=1.7in]{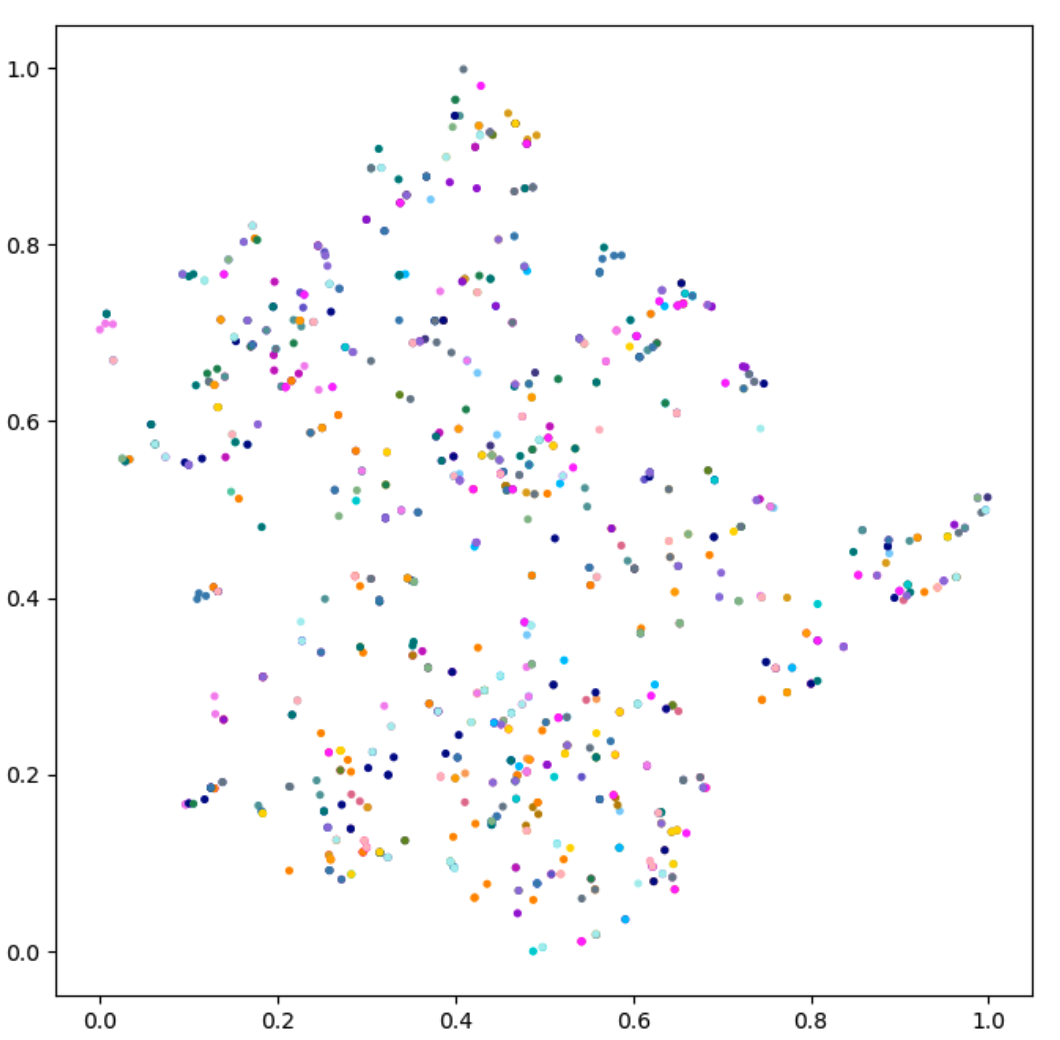}%
\label{fig_second_case}}
\caption{\textbf{A t-SNE visual comparison of the global feature} extracted by the trained discriminator from the initial data in the expert buffer and the data in the expert buffer at $2\times10^{6}$ simulation steps for \textbf{MoveBucket} task using our \textbf{Self-Imitation Learning from Demonstrations} method. (a) initial data in the expert buffer. (b) data in the expert buffer at $2\times10^{6}$ simulation steps which are mainly generated by our own policy during training. Each color represents the data of an instance.}
\label{fig_tsne}
\vspace{-4mm}
\end{figure}

With the progressive growth of the discriminator, the reward from GAIL will not diminish but we observe that the 
expert reward from GAIL still tends to decay as the training goes (see Fig. \ref{fig:curve}). This means the discriminator can be more and more successful to tell apart our generated trajectories from 
the
expert demonstration trajectories. One issue we identified that contributes to this problem is the clustering phenomenon in the trajectory space of expert demonstrations. The demonstration provided by ManiSkill challenge is obtained from per-instance RL training, therefore the underlying manipulation strategies can be quite different for different object instances (see Fig. \ref{fig_tsne}(a)). On the other hand, our policy network tends to learn a more universal strategy that can handle different object instances 
in a similar way (see Fig. \ref{fig_tsne}(b)), which comes by nature from the continuity of neural networks with the change in object instances. In general, expert demonstrations of different object instances may come from different sources and thus exhibit a highly non-uniform distribution in the trajectory space, which is not friendly to imitate using a single policy network. 

Inspired by Generative Adversarial Self-Imitation Learning (GASIL)~\cite{guo2018generative} and Self-Imitation learning from Demonstration (SILfD)~\cite{pshikhachev2021self}, we propose to fill the expert buffer with the successful trajectories generated by our policy network (see Fig. \ref{fig:pipeline}), which is so-called \textit{self-imitation learning}. We thus propose  Generative Adversarial Self-Imitation Learning from Demonstrations that combines GASIL, which leverages GAIL with Self-Imitation Learning but does not consider expert demonstrations, and SILfD, which does not use GAIL to generate rewards. 
More specifically, at the beginning of training, we initialize the expert buffer with expert demonstrations. During training, we select successful trajectories, 
that obtain the terminal rewards, of current policy through the interaction with the environment and update the expert buffer with those trajectories. As training goes, the self-generated trajectories will gradually occupy the expert buffer and yield a more uniform distribution of trajectory data. With this change, the 
expert reward increases significantly throughout the training process (see Fig. \ref{fig:curve}).


\subsection{Category-Level Instance-balancing Expert Buffer}
\label{sec:instance-balancing}
Lastly, we propose to evenly divide the expert buffer $\mathcal{B}_E$ into many slots $\{\mathcal{B}_E^j\}$ ($j=1,...,n$) of the same size, where $n$ is the number of training instances. Trajectories of one instance $O^j$, either from expert demonstrations or self-generated successful ones, can only occupy its corresponding slot $\mathcal{B}_E^j$, yielding our category-level instance-balancing (CLIB) expert buffer. At the beginning of the training, we initialize $\mathcal{B}_E^j$ with the expert trajectories of $O^j$ . In each iteration of GAIL, each slot $\mathcal{B}_E^j$ is updated by adding the success trajectories of $O^j$ in a FIFO  (First In First Out) manner, similar as before (see Fig. \ref{fig:pipeline}). 

The motivation of this design is to avoid bias in the expert buffer and improve generalization to novel object instances. For training our proposed generative adversarial self-imitation from demonstration, manipulation on some instances may happen to succeed very earlier than on the others and trajectories from these instances will quickly pop up in the expert buffer, leading to a highly biased distribution (see Fig. \ref{fig:data_dis}). 
These instances that succeed in the early stages are usually similar to each other, so GAIL may over-fit in this particular kind of instance, leading to difficulty in learning to manipulate other kinds of object instances. This is basically the Matthew Effect. Also, if some instances are harder than the others and need more time to be harnessed, our CLIB expert buffer will always keep their expert demonstrations until their successful manipulation arrives. This design can significantly improve the generalization of our learned policy to novel instances 
in certain categories that have large variations in instance manipulation difficulties.

\vspace{-2mm}
\subsection{Implementation Details}
\label{sec:details}
\subsubsection{Training details}
We call one step a one-time-step interaction with the environment of a single agent, and for each instance task, the max step for interaction is 200. 8 independent interactions run parallel to collect trajectories in a single iteration. For each iteration, SAC networks sample batch size of 1024 from replay buffer and update parameters for 4 times, and discriminator network also sample batch size of 1024 from both Generated Buffer and expert buffer and update parameters for 5 times. The total training step is set to $3\times10^{6}$ for all methods. Replay buffer size is set to $6\times10^{5}$. For Generative Adversarial Self-imitation Learning from Demonstration, the expert buffer size is set to $6\times10^{5}$, and for Category-Level Instance-Balancing expert buffer, each instance expert buffer size is set to $6\times10^{3}$. The learning rate for 
the
discriminator network is $5\times10^{-4}$, and the learning rate for 
the
policy network is $3\times10^{-4}$.

%
%

 
\subsubsection{Network Architecture}
We use the PointNet model to process point cloud state input for the original discriminator network and PointNet + Transformer model \cite{mu2021maniskill} for the policy and value network in the SAC part. The input of the point cloud state consists of the point cloud of the environment and the robot state. For the value and discriminator network, the action vector is concatenated to the robot state to be processed together. Then robot state with action is concatenated to each point in the point cloud. This concatenated point cloud is then processed as described in section \ref{sec:progressive} to form K+4 sub-features. 

\vspace{-3mm}

\section{Experiments}
\vspace{-1mm}
\begin{table}[]\small
\vspace{+4mm}
\caption{Statistics for Four Tasks}
\label{tasks_statics}
\centering
\scalebox{0.9}{
\begin{tabular}{@{}c|cc|c|c@{}}
\toprule
\multirow{2}{*}{Task} & \multicolumn{2}{c|}{Objects} & \multirow{2}{*}{\begin{tabular}[c]{@{}c@{}}Dual-arm\\ Robot\end{tabular}} & \multirow{2}{*}{\begin{tabular}[c]{@{}c@{}}Action Space\\ Dimension\end{tabular}} \\ \cmidrule(lr){2-3}
                      & Train         & Test         &                                                                                   &                                                                                   \\ \midrule
OpenCabinetDrawer     & 42            & 10           & No                                                                                & 13                                                                                \\
OpenCabinetDoor       & 25            & 10           & No                                                                                & 13                                                                                \\
PushChair             & 26            & 10           & Yes                                                                               & 22                                                                                \\
MoveBucket            & 29            & 10           & Yes                                                                               & 22                                                                                \\ \bottomrule
\end{tabular}}
\vspace{-4mm}
\end{table}

\subsection{Benchmark}
\label{sec:benchmark}
We evaluate our methods on SAPIEN \cite{xiang2020sapien} Manipulation Skill Benchmark \cite{mu2021maniskill} (ManiSkill). This section will briefly introduce the task settings.
\subsubsection{Task Introduction}
There are four tasks in ManiSkill and these tasks cover different types of object motions. Statistics for the four tasks are summarized in Table \ref{tasks_statics}.
\begin{itemize} 

\item \textbf{OpenCabinetDrawer}: In this task, a single-arm robot is required to open a designed drawer on a cabinet.
\item \textbf{OpenCabinetDoor}: In this task, a single-arm robot is required to open a designed door on a cabinet.
\item \textbf{PushChair}: In this task, a dual-arm robot is required to push a swivel chair to a target location on the ground and prevent it from falling over.
\item \textbf{MoveBucket}: In this task, a dual-arm robot is required to move a bucket with a ball inside and lift it to a platform.

\end{itemize}

\subsubsection{Observation}
The observation of the task is composed of two components: (\romannumeral1) robot state (a vector that describes the agent's state, including pose, velocity, the angular velocity of the moving platform of the robot, joint angles, and joint velocities of all robot joints, positions, and velocities of the robot fingers) ; (\romannumeral2) point cloud of the scene(6$+$k dimensions: 3 XYZ positions for each point, 3 RGB values for each point, k task-relevant segmentation masks).

\subsubsection{Dense Reward}
ManiSkill provides carefully designed dense reward functions for each task. Since the dense reward function is expensive to design and is task-specific and needs to be manually adjusted for the particular task, we don't assume we have it in the main experiments. However, we provide an evaluation of our methods with an additional handcrafted dense reward section \ref{sec:dense_reward}.
\vspace{-1mm}
\subsection{Results and Analysis}
\label{sec:main_results}
\vspace{-1mm}
\begin{table*}[]
\vspace{+2mm}
\caption{Main Results and Ablation Studies on the four tasks of ManiSkill Challenge Benchmark}
\label{main_results}
\centering
\scalebox{0.9}{
\begin{tabular}{@{}c|c|c|c|c|cccccc@{}}
\toprule
Method             & GAIL                 & \begin{tabular}[c]{@{}c@{}}Progressive\\ Growing of\\ Discriminator\end{tabular} & \begin{tabular}[c]{@{}c@{}}Self-Imitation\\ Learning from\\ Demonstrations\end{tabular} & \begin{tabular}[c]{@{}c@{}}CLIB\\ Expert Buffer\end{tabular} & Task  & \begin{tabular}[c]{@{}c@{}}OpenCabinet\\ Drawer\end{tabular} & \begin{tabular}[c]{@{}c@{}}OpenCabinet\\ Door\end{tabular} & PushChair & MoveBucket & Avg       \\ \midrule
\multirow{2}{*}{\uppercase\expandafter{\romannumeral1}} & \multirow{2}{*}{\checkmark} & \multirow{2}{*}{}                                                                       & \multirow{2}{*}{}                                                                & \multirow{2}{*}{}                                                                 & Train & 0.43$\pm$0.03                                                    & 0.25$\pm$0.03                                                  & 0.19$\pm$0.02 & 0.16$\pm$0.02  & 0.26$\pm$0.03 \\
                   &                      &                                                                                         &                                                                                  &                                                                                   & Val   & 0.36$\pm$0.03                                                    & 0.15$\pm$0.03                                                  & 0.11$\pm$0.01 & 0.11$\pm$0.02  & 0.18$\pm$0.03 \\ \midrule
\multirow{2}{*}{\uppercase\expandafter{\romannumeral2}} & \multirow{2}{*}{\checkmark} & \multirow{2}{*}{\checkmark}                                                                       & \multirow{2}{*}{}                                                             & \multirow{2}{*}{}                                                                 & Train & 0.52$\pm$0.03                                                    & 0.38$\pm$0.06                                                  & 0.21$\pm$0.03 & 0.19$\pm$0.03  & 0.33$\pm$0.03 \\
                   &                      &                                                                                         &                                                                                  &                                                                                   & Val   & 0.49$\pm$0.04                                                    & 0.27$\pm$0.04                                                  & 0.16$\pm$0.04 & 0.16$\pm$0.03  & 0.27$\pm$0.04 \\ \midrule
\multirow{2}{*}{\uppercase\expandafter{\romannumeral3}} & \multirow{2}{*}{\checkmark} & \multirow{2}{*}{}                                                                    & \multirow{2}{*}{\checkmark}                                                                & \multirow{2}{*}{}                                                                 & Train & 0.45$\pm$0.02                                                    & 0.25$\pm$0.05                                                  & 0.21$\pm$0.03 & 0.17$\pm$0.02  & 0.27$\pm$0.02 \\
                   &                      &                                                                                         &                                                                                  &                                                                                   & Val   & 0.41$\pm$0.06                                                    & 0.24$\pm$0.03                                                  & 0.16$\pm$0.02 & 0.14$\pm$0.03  & 0.25$\pm$0.02 \\ \midrule
\multirow{2}{*}{\uppercase\expandafter{\romannumeral4}} & \multirow{2}{*}{\checkmark} & \multirow{2}{*}{}                                                                    & \multirow{2}{*}{\checkmark}                                                                & \multirow{2}{*}{\checkmark}                                                              & Train & 0.58$\pm$0.05                                                    & 0.37$\pm$0.05                                                  & 0.23$\pm$0.05 & 0.25$\pm$0.05  & 0.36$\pm$0.03 \\
                   &                      &                                                                                         &                                                                                  &                                                                                   & Val   & 0.62$\pm$0.04                                                    & 0.30$\pm$0.04                                                  & 0.18$\pm$0.04 & 0.16$\pm$0.03  & 0.31$\pm$0.03 \\ \midrule
\multirow{2}{*}{\uppercase\expandafter{\romannumeral5}} & \multirow{2}{*}{\checkmark} & \multirow{2}{*}{\checkmark}                                                                    & \multirow{2}{*}{\checkmark}                                                             & \multirow{2}{*}{\checkmark}                                                              & Train & \textbf{0.61$\pm$0.04}                                                    & \textbf{0.41$\pm$0.05}                                                  & \textbf{0.27$\pm$0.04} & \textbf{0.27$\pm$0.04}  & \textbf{0.39$\pm$0.03} \\
                   &                      &                                                                                         &                                                                                  &                                                                                   & Val   & \textbf{0.65$\pm$0.04}                                                    & \textbf{0.33$\pm$0.05}                                                  & \textbf{0.24$\pm$0.03} & \textbf{0.19$\pm$0.04}  & \textbf{0.36$\pm$0.04} \\ \bottomrule
\end{tabular}}
\vspace{-2mm}
\end{table*}

The results are summarised in Table \ref{main_results}. We evaluated our methods on four tasks via 100 trials with three different random seeds. In Table \ref{main_results}, with [Progressive Growing of Discriminator], [Self-Imitation Learning from Demonstrations] and [CLIB Expert Buffer], our (Method \uppercase\expandafter{\romannumeral5}) outperforms [GAIL] (Method \uppercase\expandafter{\romannumeral1}) by \textbf{13\%} and \textbf{18\%} averaged across four tasks on training and validation sets. In the following, we will analyze the contribution from each individual technique we introduce.

\subsubsection{Effect of the Progressive Growing of Discriminator on top of GAIL}From Table \ref{main_results} we can see that compared with [GAIL] (Method \uppercase\expandafter{\romannumeral1}), [Progressive Growing of Discriminator](Method \uppercase\expandafter{\romannumeral2}) significantly improves success rate by \textbf{7\%} and \textbf{9\%} averaged across four tasks on train and validation sets, respectively. Fig. \ref{fig:curve} shows the expert reward curve from the discriminator during training. With a strong discriminator, the expert reward from the discriminator in GAIL will quickly drop to a very low level since the discriminator will quickly learn to distinguish between the expert data and the data generated by the RL part. This will have a significant negative impact on RL learning, which is largely dependent on the reward's guidance. However, when the discriminator grows from a simpler network to a more sophisticated network, the reward provided by the discriminator is more stable. By progressive growth, the discriminator can be shaped as a more complex structure to produce more comprehensive rewards, which will, in turn, boost the generalization ability of RL. Moreover, by comparing the method of progressively growing both the generator and the discriminator with the method of only progressively growing the discriminator, we find that with the growing generator, the method will decrease by \textbf{17\%} and \textbf{15\%} averaged across four tasks on training and validation sets (Table \ref{ablation_results}), respectively. We consider that different from PG-GAN~\cite{karras2018progressive} which increases the generation resolution from coarse to fine, the output of our generator is always in the same action space so we only need a strong generator and gradually increase the sophistication of the discriminator.

\begin{table}[]\tiny
\caption{Ablation Studies on Progressive Growing}
\label{ablation_results}
\centering
\begin{tabular}{c|cccccc}
\hline
\begin{tabular}[c]{@{}c@{}}Progressive\\ Growing\end{tabular}                            & Task  & \begin{tabular}[c]{@{}c@{}}OpenCabinet\\ Drawer\end{tabular} & \begin{tabular}[c]{@{}c@{}}OpenCabinet\\ Door\end{tabular} & PushChair  & MoveBucket & Avg        \\ \hline
\multirow{2}{*}{Discriminator}                                                           & Train & \textbf{0.52$\pm$0.03}                                                   & \textbf{0.38$\pm$0.06}                                                 & \textbf{0.21$\pm$0.03} & \textbf{0.19$\pm$0.03} & \textbf{0.33$\pm$0.03} \\
                                                                                         & Val   & \textbf{0.49$\pm$0.04}                                                   & \textbf{0.27$\pm$0.04}                                                 & \textbf{0.16$\pm$0.04} & \textbf{0.16$\pm$0.03} & \textbf{0.27$\pm$0.04} \\ \hline
\multirow{2}{*}{\begin{tabular}[c]{@{}c@{}}Discriminator\\ and Generator\end{tabular}} & Train & 0.25$\pm$0.03                                                   & 0.19$\pm$0.03                                                 & 0.11$\pm$0.04 & 0.10$\pm$0.02 & 0.16$\pm$0.02 \\
                                                                                         & Val   & 0.16$\pm$0.03                                                   & 0.16$\pm$0.04                                                 & 0.10$\pm$0.02 & 0.06$\pm$0.01 & 0.12$\pm$0.02 \\ \hline
\end{tabular}
\vspace{-3mm}
\end{table}

\subsubsection{Effect of the Self-Imitation Learning from Demonstrations on top of GAIL} From Table \ref{main_results} we can see that 
[GAIL] plus [Self-Imitation Learning from Demonstrations](Method \uppercase\expandafter{\romannumeral3}) outperforms [GAIL](Method \uppercase\expandafter{\romannumeral1}) by \textbf{7\%} on the validation sets of four tasks on average.
Since the [Self-Imitation Learning from Demonstrations](Method \uppercase\expandafter{\romannumeral3}) updates the expert buffer with the successful trajectories generated by the recent SAC policy, the trajectories in the expert buffer will resemble the generated trajectories. This will let the reward from the discriminator converge to a relatively high value, as shown in Fig. \ref{fig:curve}. Instead of only imitating from initial expert demonstrations which are circumscribed by the difference between each instance, Self-Imitation Learning provides the expert buffer with more various trajectories. So, the policy from (Method \uppercase\expandafter{\romannumeral3}) will not be constrained by the limitations of expert demonstrations, thus the generalization ability is improved. Fig. \ref{fig_tsne} shows a visual comparison of features using t-SNE between the initial data in the expert buffer and the data in the expert buffer after certain epochs of training. The initial expert data is generated by different single-instance-specialized RL agents on different instances and thus has many separate manifolds in the feature space.
It is hard for a continuous policy network with respect to the object point cloud inputs to imitate these demonstrations that are discontinuous.
On the other hand, the manifold of trajectories generated by our generalized policy is more smooth and continuous and has a well-mixed distribution.
\subsubsection{Effect of CLIB Expert Buffer}
\begin{figure}[!t]
\vspace{+3mm}
\centering
\includegraphics[width=3.2in]{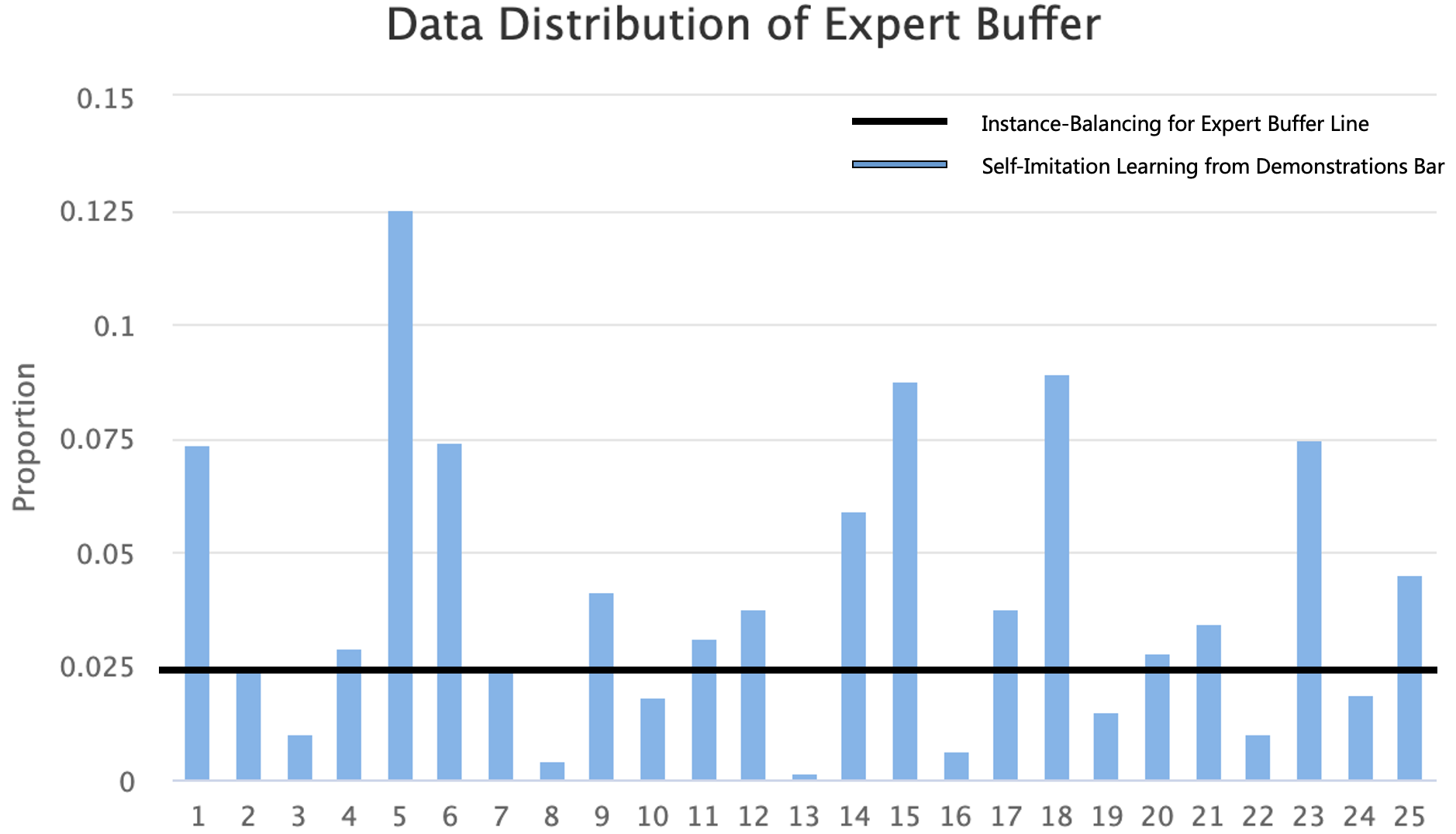}
\caption{\textbf{The data distribution of expert buffer on different instances} at $2\times10^{6}$ simulation steps for \textbf{OpenCabinetDrawer} task. The black line represents the data distribution of the expert buffer using our \textbf{CLIB Expert Buffer} method which is evenly distributed on different instances. The blue bar represents the data distribution of the expert buffer without using this method. The X-axis and Y-axis represent different instances and the proportion of their data in the expert buffer.}
\label{fig:data_dis}
\vspace{-3mm}
\end{figure}

\begin{figure}[!t]
\centering
\includegraphics[width=3.0in]{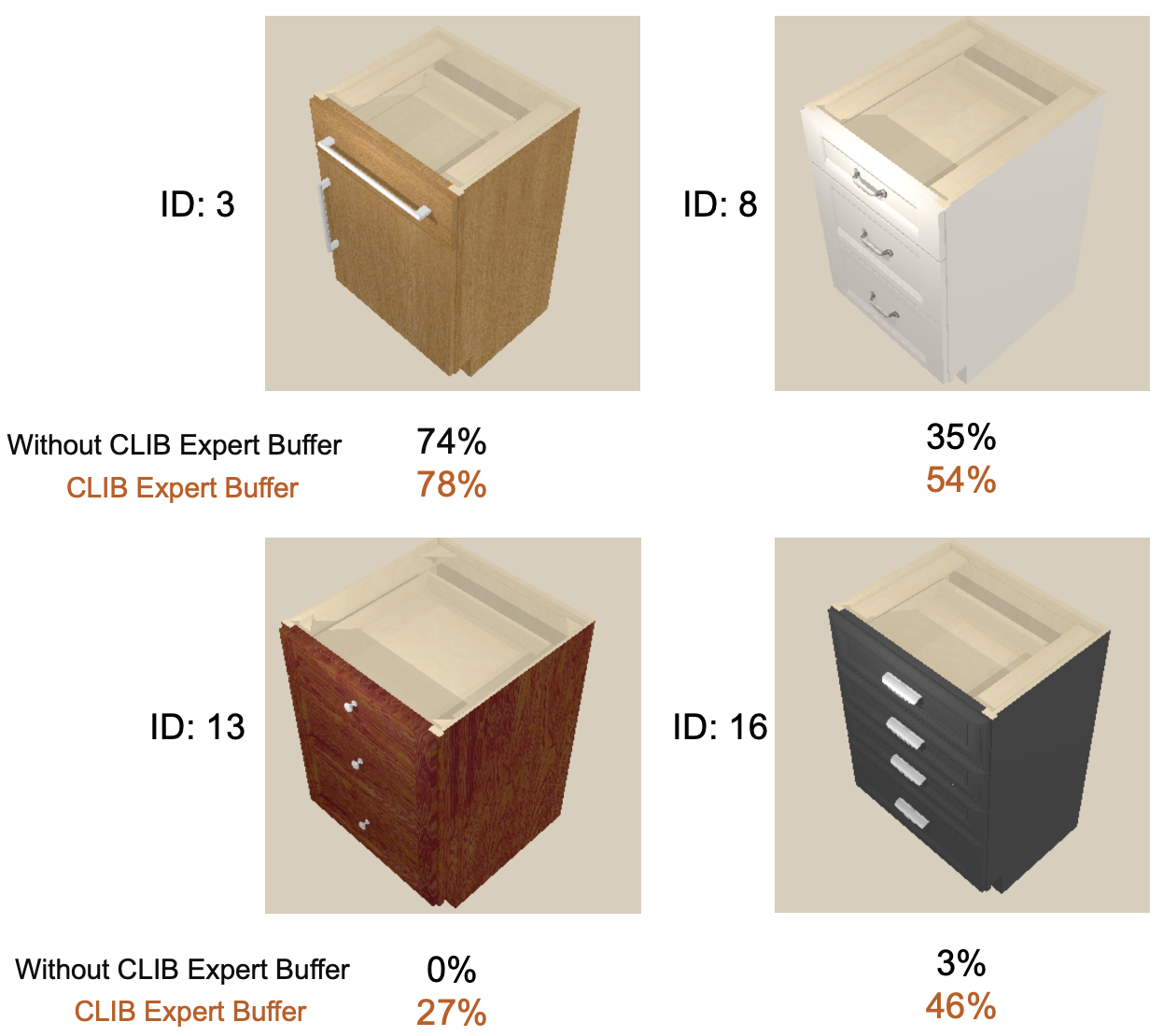}
\caption{\textbf{The success rate comparison on some variant instances before and after using our CLIB expert buffer}. In \textbf{OpenCabinetDrawer} task, the handles of cabinets have many variations, and some handles, for example, nail shaped, are hard to be opened (ID:13). Without CLIB, the Expert Buffer will soon forget the initial demonstrations for these instances and the agent will not learn to finish these tasks.}
\label{fig:CLIB_drawer}
\vspace{-4mm}
\end{figure}

From Table \ref{main_results} we can observe that by making the expert buffer balanced during training, our [CLIB Expert Buffer](Method \uppercase\expandafter{\romannumeral4}) method surpasses [Self-Imitation Learning from Demonstrations](Method \uppercase\expandafter{\romannumeral3}) by \textbf{6\%} on the validation sets of four tasks on average. Besides, the improvement is significant when the instances vary greatly in size, shapes, etc. like cabinet drawers in \textbf{OpenCabinetDrawer} task. Fig. \ref{fig:data_dis} shows the distribution of data in the expert buffer on different instances in the later stage of training. It shows that during training, our [CLIB Expert Buffer](Method \uppercase\expandafter{\romannumeral4}) method ensures that the data in the expert buffer is evenly distributed on instances while the expert data is much unevenly distributed on instances without using the method. This means the successful trajectories in the expert buffer on some instances may be less and less or even there is no successful trajectory on these instances if the RL hasn’t generated successful trajectories on these instances in the early time of training. Thus, the RL may lose the guidance of expert data on these instances. An example of loss of guidance is shown in Fig. \ref{fig:CLIB_drawer}. Some instances of the drawers with handles that would not easily be grasped by paralleled gripper (ID: 13 \& 16) are
hard to be manipulated successfully. Without Category-Level Instance-Balancing (CLIB), these instances' 
demonstrations
will soon be ejected out of the expert buffer, so the policy network will have great difficulties in learning how to operate these drawers. However, with the help of CLIB, the expert buffer always reserves demonstrations for these drawers 
so the agent also gets some successful knowledge on these instances.

\begin{figure}[!t]
\vspace{+2mm}
\centering
\includegraphics[width=3.3in]{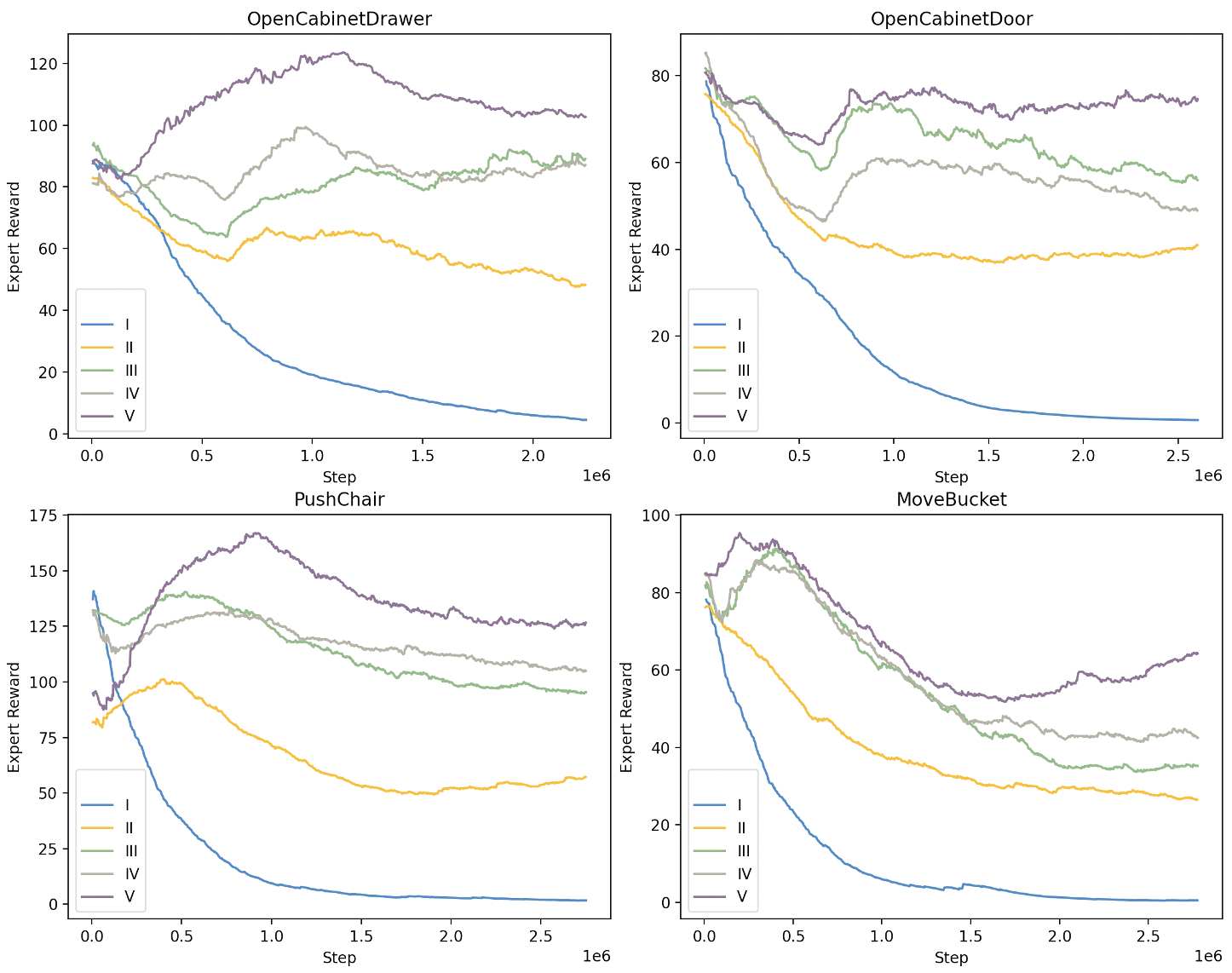}
\caption{\textbf{The expert reward curve during training.} The expert reward comes from the discriminator. When the expert reward is relatively small, it indicates that the discriminator can easily distinguish between the expert data and the data generated by the policy. Notice that our method (curve \uppercase\expandafter{\romannumeral5}) always achieves the highest value.}
\label{fig:curve}
\vspace{-2mm}
\end{figure}

Finally, we combine these methods together. Fig. \ref{fig:curve} reveals that when using Self-Imitation Learning from Demonstrations, the reward is higher and stabler than the method that only uses progressive growing of discriminator. This occurs since Self-Imitation Learning from Demonstrations will update expert buffer with self policy generated trajectories, which adds difficulties to discriminate between newly generated trajectories and those stored in the expert buffer. Therefore the expert reward remains higher in Self-Imitation Learning from Demonstrations methods. However, the structure of the discriminator does not change in GAIL with Self-Imitation Learning from Demonstration, the discriminator will be misled and gradually loses its discriminative power. The combination of Self-Imitation Learning from Demonstrations, CLIB Expert Buffer, and progressive growing of discriminator can help to solve this problem. Our final method reaches the highest success rate on both training set and validation set, improving \textbf{13\%} success rate on average training and \textbf{18\%} success rate on average validation.

\subsection{More Results with Handcrafted Dense Reward}
\label{sec:dense_reward}

\begin{table}[]\tiny
\vspace{+4mm}
\caption{Additional Experiments with Dense Reward}
\label{reward_results}
\centering
\begin{tabular}{@{}c|lccccc@{}}
\toprule
Method                                                                        & Task  & \begin{tabular}[c]{@{}c@{}}OpenCabinet\\ Drawer\end{tabular} & \begin{tabular}[c]{@{}c@{}}OpenCabinet\\ Door\end{tabular} & PushChair  & MoveBucket & Avg        \\ \midrule
\multirow{2}{*}{SAC}                                                          & Train & 0.93$\pm$0.02                                                   & 0.69$\pm$0.03                                                 & 0.52$\pm$0.02 & \textbf{0.76$\pm$0.04} & 0.72$\pm$0.04 \\
                                                                              & Val   & 0.92$\pm$0.02                                                   & 0.35$\pm$0.03                                                 & 0.35$\pm$0.04 & 0.61$\pm$0.04 & 0.56$\pm$0.03 \\ \midrule
\multirow{2}{*}{\begin{tabular}[c]{@{}c@{}}GAIL +\\ Dense Reward\end{tabular}} & Train & 0.97$\pm$0.02                                                   & 0.92$\pm$0.02                                                 & 0.54$\pm$0.03 & 0.50$\pm$0.03 & 0.73$\pm$0.04 \\
                                                                              & Val   & 0.88$\pm$0.01                                                   & \textbf{0.78$\pm$0.02}                                                 & 0.42$\pm$0.04 & 0.47$\pm$0.04 & 0.63$\pm$0.04 \\ \midrule
\multirow{2}{*}{\begin{tabular}[c]{@{}c@{}}\uppercase\expandafter{\romannumeral5} +\\ Dense Reward\end{tabular}}    & Train & \textbf{0.99$\pm$0.01}                                                   & \textbf{0.94$\pm$0.02}                                                 & \textbf{0.61$\pm$0.04} & 0.64$\pm$0.05 & \textbf{0.80$\pm$0.04} \\
                                                                              & Val   & \textbf{0.96$\pm$0.01}                                                   & 0.76$\pm$0.04                                                 & \textbf{0.46$\pm$0.04} & \textbf{0.63$\pm$0.04} & \textbf{0.70$\pm$0.05} \\ \bottomrule
\end{tabular}
\vspace{-4mm}
\end{table}
Additionally, we evaluate our methods with additional handcrafted dense rewards provided by ManiSkill. The results are summarised in Table \ref{reward_results}.
We linearly add the expert reward from the discriminator and the environment reward together and use this total reward to update the policy and value networks in RL. More specifically, we assume $r_t = -\log D_\phi(S_t,a_t) + \alpha*r_{env}$, where $r_{env}$ is provided by ManiSkill Challenge and $\alpha$ is 0.5 in our experiment.
We find that our methods with additional dense reward, which ranks first place on the ``no external annotation'' track of ManiSkill Challenge 2021, can outperform the GAIL+SAC baseline by \textbf{7\%} averaged across four tasks on both training and validation sets.

\vspace{-1mm}
\subsection{Limitations}
The limitation of applying our methods without any dense reward is revealed in long-horizon tasks. For example, in tasks of \text{MoveBucket}, these tasks require 
robots
to achieve certain states at each stage (\text{e.g. firmly grasp the bucket handle before moving it}). Without the guidance from an accurate stage-wise reward function, the overall success rate is not yet perfect. However, our contributions are orthogonal to reward engineering. It is very promising to combine our method with well-shaped dense reward, yielding excellent performance on the tasks.

\section{CONCLUSIONS}
\vspace{-1mm}
In this paper, we, for the first time, tackle the problem of category-level object manipulation via generative adversarial self-imitation learning from demonstrations. We build our method upon GAIL with SAC. We propose several important techniques to improve the baseline, including combining GAIL with self-imitation learning from demonstrations, progressive growing of discriminator, and instance-balancing for expert buffer. Our experiments shows that our methods can reach a much higher success rate on four tasks from ManiSkill benchmark than existing baselines and our ablation studies further validates the contribution of each technique.








\addtolength{\textheight}{-12cm}   

\bibliographystyle{IEEEtran}
\end{document}